\documentclass[runningheads]{llncs}
\usepackage{algorithm}  
\usepackage[noend]{algpseudocode}
\usepackage{amsmath,amsfonts,amssymb}
\usepackage{bm}
\usepackage{caption}
\usepackage{float}
\usepackage{graphicx}
\usepackage[misc]{ifsym}
\usepackage{makecell}
\usepackage{pifont}
\usepackage{subcaption}
\newcommand{\cmark}{\ding{51}}%
\newcommand{\xmark}{\ding{55}}%

\begin{document}
\title{Federated Contrastive Learning for Decentralized Unlabeled Medical Images}
%
%
\author{Nanqing Dong \and Irina Voiculescu}
%
\authorrunning{Dong et al.}
%
\institute{Department of Computer Science, University of Oxford, Oxford, UK\\
\email{nanqing.dong@cs.ox.ac.uk}}

\maketitle              
\begin{abstract}
A label-efficient paradigm in computer vision is based on self-supervised contrastive pre-training on unlabeled data followed by fine-tuning with a small number of labels. Making practical use of a federated computing environment in the clinical domain and learning on medical images poses specific challenges. In this work, we propose \texttt{FedMoCo}, a robust federated contrastive learning (FCL) framework, which makes efficient use of decentralized unlabeled medical data. \texttt{FedMoCo} has two novel modules: \textit{metadata transfer}, an inter-node statistical data augmentation module, and \textit{self-adaptive aggregation}, an aggregation module based on representational similarity analysis. To the best of our knowledge, this is the first FCL work on medical images. Our experiments show that \texttt{FedMoCo} can consistently outperform \texttt{FedAvg}, a seminal federated learning framework, in extracting meaningful representations for downstream tasks. We further show that \texttt{FedMoCo} can substantially reduce the amount of labeled data required in a downstream task, such as COVID-19 detection, to achieve a reasonable performance.
\end{abstract}
\section{Introduction}
Recent studies in self-supervised learning (SSL)~\cite{de1994learning} have led to a renaissance of research on contrastive learning (CL)~\cite{baldi1991contrastive}. Self-supervised or unsupervised CL aims to learn transferable representations from unlabeled data. In a CL framework, a model is first pre-trained on unlabeled data in a self-supervised fashion via a contrastive loss, and then fine-tuned on labeled data. Utilizing the state-of-the-art (SOTA) CL frameworks~\cite{he2020momentum,misra2020self,chen2020simple,tian2020makes}, a model trained with only unlabeled data plus a small amount of labeled data can achieve comparable performance with the same model trained with a large amount of labeled data on various downstream tasks.

As a data-driven approach, deep learning has fueled many breakthroughs in medical image analysis (MIA). Meanwhile, large-scale fully labeled medical datasets require considerable human annotation cost, which makes data scarcity a major bottleneck in practical research and applications. To leverage unlabeled data, CL \cite{he2020sample,chen2020momentum,sowrirajan2020moco} has obtained promising results on unlabeled data, yet none of these studies consider federated learning (FL), which can handle sensitive data stored in multiple devices \cite{li2019privacy,rieke2020future,sheller2020federated,dou2021federated}. Data privacy regulations restrict collecting clinical data from different hospitals for conventional data-centralized CL. Under FL protocols, medical images (either raw or encoded) must not be exchanged between data nodes. In this work, we design an FCL framework on decentralized medical data (i.e.~medical images stored on multiple devices or at multiple locations) to extract useful representations for MIA.

There are two direct \textit{negative} impacts of the FL environment on CL. First, different imaging protocols in hospitals (nodes) create domain shift~\cite{dong2018unsupervised}. In contrast to CL on centralized data, each node in FCL only has access to its local data, which has a smaller variation in sample distribution. This impairs the CL performance on single nodes. Second, without the supervision of ground truth labels, there is no guarantee for the performance of CL in each node. Applying supervised FL frameworks on CL directly might lead to worse generalization ability and hence poorer outcome as the performance of FCL could be dominated by the performance of a few nodes. How to aggregate CL models across nodes is still an open question.

\begin{figure}[t!]
    \centering
    \includegraphics[width=0.75\textwidth]{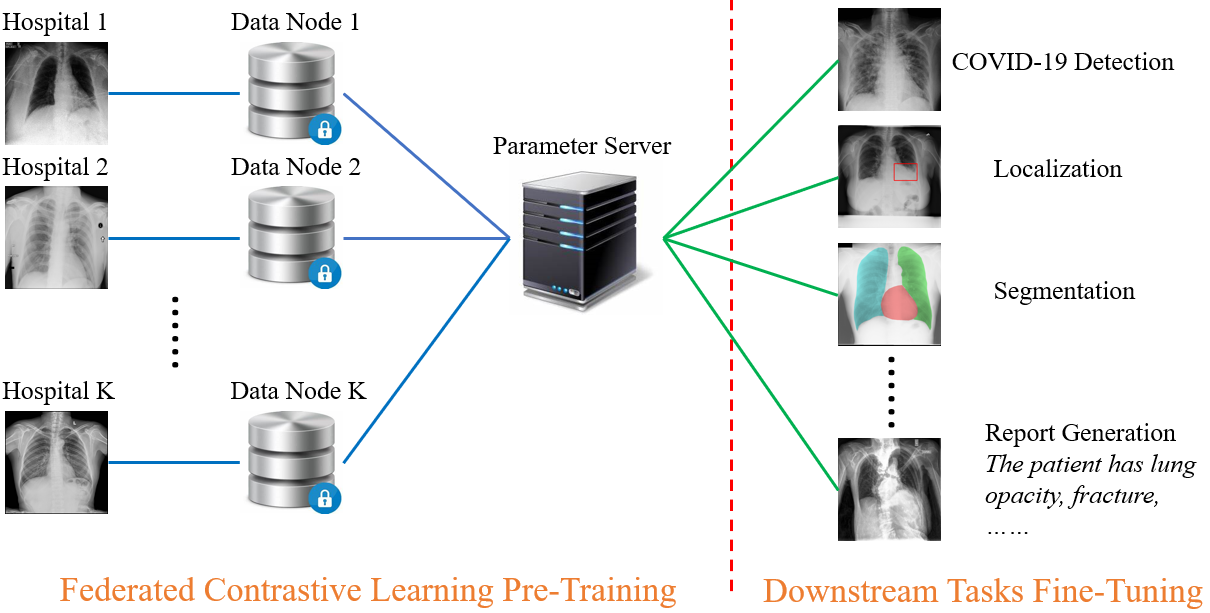}
\caption{Illustration of FCL workflow. Each node works on an independent local dataset. The parameter server works with data nodes for periodic synchronization, metadata transfer (Sec.~\ref{sec:inter}), and self-adaptive aggregation (Sec.~\ref{sec:agg}). For each downstream task, the pre-trained model weights are fine-tuned with small amount of labeled data in either supervised or semi-supervised fashion. Due to data privacy, the labeled data of a downstream task can only be accessed {\em locally}.}
\label{fig:fed}
\end{figure}

We introduce \texttt{FedMoCo}, a robust FCL framework for MIA (see Fig.~\ref{fig:fed}). \texttt{FedMoCo} uses MoCo \cite{he2020momentum} as the intra-node CL model. To mitigate the above hurdles, we propose \textit{metadata transfer}, an inter-node augmentation module utilizing Box-Cox power transformation \cite{box1964analysis} and Gaussian modeling; we also propose \textit{self-adaptive aggregation}, a module based on representational similarity analysis (RSA) \cite{kriegeskorte2008representational}. We empirically evaluate \texttt{FedMoCo} under various simulated scenarios. Our experiments show that \texttt{FedMoCo} can consistently outperform \texttt{FedAvg} \cite{mcmahan2017communication}, a SOTA FL framework; \texttt{FedMoCo} can efficiently reduce annotation cost in downstream tasks, such as COVID-19 detection. By pre-training on unlabeled non-COVID datasets, \texttt{FedMoCo} requires only $3\%$ examples of a labeled COVID-19 dataset to achieve $90\%$ accuracy. Our contributions are threefold: (1) to the best of our knowledge, this is the first work of FCL on medical images; (2) we propose \texttt{FedMoCo} with two novel modules; (3) our results provide insights into future research on FCL.

\section{Method}
\noindent\textbf{Problem Formulation.}
$K >1$ denotes the number of nodes and $\mathcal{D}_k$ is an {\em unlabeled\/} dataset stored in node $k \le K$. $x_i$ is a medical image in $\mathcal{D}_k = \{x_i\}_{i=1}^{n_k}$, with $n_k = |\mathcal{D}_k|$. As is standard practice in FL, $\mathcal{D}_k \cap \mathcal{D}_l = \emptyset$ for $k \neq l$ and $\{\mathcal{D}_k\}_{k=1}^K$ are non-IID data. There is an additional master node, which does not store any clinical data. The master node is implemented as a \textit{parameter server} (PS) \cite{li2014communication}. 

The model of interest is $f_{\theta}$ and its parameter set $\theta_0$ is randomly initialized in PS. At the beginning of the federated training, $K$ copies of $\theta_0$ are distributed into each node as $\{\theta_k\}_{k=1}^K$, i.e.~we fully synchronize the data nodes with PS. An important FL protocol is data privacy preservation: exchanging training data between nodes is strictly prohibited. Instead, node $k$ updates $\theta_k$ by training on $\mathcal{D}_k$ independently. After the same number of local epochs, $\{\theta_k\}_{k=1}^K$ are aggregated into $\theta_0$ in PS. Again, we synchronize $\{\theta_k\}_{k=1}^K$ with aggregated $\theta_0$. This process is repeated until certain criteria are met. To enforce data privacy, we only ever exchange model parameters $\{\theta_k\}_{k=0}^K$ and metadata between the data nodes and PS. The learning outcome is $\theta_0$ for relevant downstream tasks.

\subsection{Intra-Node Contrastive Learning}
\label{sec:intra}
We create a positive pair by generating two random views of the same image through data augmentation. A negative pair is formed by taking two random views from two different images. Given an image $x$ and a family of stochastic image transformations $\mathcal{T}$, we randomly sample two transformations $\tau$ and $\tau'$ to have two random views $\tau(x)$ and $\tau'(x)$ which form a positive pair. Let $z$ denote the representation of an image extracted by a CNN encoder $f_{\theta}$. In contrast to previous works \cite{wu2018unsupervised,he2020momentum,chen2020simple}, we add a ReLU function between the last fully-connected layer and a L2-normalization layer, which projects all extracted features into a non-negative feature space. The non-negative feature space is a prerequisite for Sec.~\ref{sec:inter}. For query image $x\in \mathcal{D}_k$, we have positive pair $z_q = f_{\theta_k^{q}}(\tau(x))$ and $z_0 = f_{\theta_k^{d}}(\tau'(x))$. If there are $N$ negative examples $\{x_i\}_{i=1}^N$, the contrastive loss InfoNCE \cite{oord2018representation} is defined as 
\begin{equation}
\mathcal{L}_q = - \log \frac{\exp(z_q \cdot z_0 / te)}{\sum_{i=0}^{N} \exp(z_q \cdot z_i / te)},
\label{eq:nce}
\end{equation}
where the temperature parameter is $te$.

We use the dynamic dictionary with momentum update in MoCo \cite{he2020momentum} to maintain a large number of negative examples in Eq.~\ref{eq:nce}. In node $k$, there are two CNNs, a CNN encoder $(\theta_k^{q})$ for the query image and another $(\theta_k^{d})$ for the corresponding positive example which are defined as 
\begin{equation}
    \theta_k^{q} ~\leftarrow~ \theta_k, ~~~~ \theta_k^{d} ~\leftarrow~ m \theta_k^{d} + (1 - m)\theta_k^{q},
    \label{eq:moment}
\end{equation}
where $m \in [0, 1)$ is the momentum coefficient. For query image $x$, we have $z_q = f_{\theta_k^{q}}(\tau(x))$ and $z_0 = f_{\theta_k^{d}}(\tau'(x))$.

\subsection{Metadata Transfer}
\label{sec:inter}
The dynamic dictionary of node $k$ must not exchange information with other nodes. This limits the sample variation of the dynamic dictionary, which limits CL performance. In addition, a node overfitted to local data may not generalize well to other nodes. We overcome these two hurdles by utilizing metadata from the encoded feature vectors in each node. For node $k$, after the full synchronization with PS (i.e.~before the start of next round of local updates), we extract feature vectors $\{f_{\theta_k}(x_i)\}_{i=1}^{n_k}$ of $\mathcal{D}_k$.  In order to enforce a Gaussian-like distribution of the features we use the Box-Cox power transformation (BC). BC is a reversible transformation defined as 
\begin{equation}
    BC(x) = 
    \begin{cases}
        \frac{x^{\lambda} - 1}{\lambda}, & \lambda \neq 0\\
        \log(x), & \lambda = 0
    \end{cases}
    \label{eq:bc}
\end{equation}
where $\lambda$ controls the skewness of the transformed distribution. Then, we calculate the mean and covariance\footnote{For large-scale datasets, the mean and covariance could be estimated by using random samples from the population. Here, we use the notation $n_k$ for simplicity.} of the transformed features 
\begin{equation}
    \bm{\mu}_k = \frac{\sum_{i=1}^{n_k} y_i}{n_k}, ~~~~ \bm{\Sigma}_k = \frac{\sum_{i=1}^{n_k} (y_i - \bm{\mu}_k)(y_i - \bm{\mu}_k)^{\bm{T}}}{n_k - 1}
    \label{eq:meta}
\end{equation}
where $y_i = \mathrm{BC}(f_{\theta_k}(x_i))$. The metadata of the learned representations in each node are collected by PS as $\{(\bm{\mu}_k, \bm{\Sigma}_k)\}_{k=1}^K$. $\{(\bm{\mu}_j, \bm{\Sigma}_j)\}_{j \neq k}$ is sent to node $k$. We name this operation as \textit{metadata transfer}. Metadata transfer improves the CL performance of node $k$ by augmenting the dynamic dictionary of node $k$ statistically with metadata collected from all the other nodes $j \neq k$.

For each pair of $(\bm{\mu}_k, \bm{\Sigma}_k)$, there is a corresponding Gaussian distribution $\mathcal{N}(\bm{\mu}_k, \bm{\Sigma}_k)$. In the next round of local updates, we increase the sample variation in each node by sampling new points from the Gaussian distributions when minimizing Eq.~\ref{eq:nce}. Specifically, for node $k$, we increase the number of negative examples in Eq.~\ref{eq:nce} from $N$ to $\left \lfloor N(1 + \eta) \right \rfloor$, where $\eta \ge 0$ is a hyper-parameter to control the level of interaction between the node $k$ and the other nodes (e.g.~$\eta = 0$ means no interaction). We sample $\left \lfloor \frac{\eta N}{K - 1} \right \rfloor$ examples from each $\mathcal{N}(\bm{\mu}_l, \bm{\Sigma}_l) \ \forall \ l \neq k$. Let $\tilde{y} \sim \mathcal{N}(\bm{\mu}_l, \bm{\Sigma}_l) \ \forall \ l \neq k$, we have $\tilde{z} = BC^{-1}(\tilde{y})$ and the new contrastive loss is 
\begin{equation}
    \mathcal{L}_q = - \log \frac{\exp(z_q \cdot z_0 / \tau)}{\sum_{i=0}^{N} \exp(z_q \cdot z_i / \tau) + \sum_{j=1}^{(K-1) \left \lfloor \frac{\eta N}{K - 1} \right \rfloor} \exp(z_q \cdot \tilde{z_j}/ \tau)}
\label{eq:meta_trans}
\end{equation}

\subsection{Self-Adaptive Aggregation}
\label{sec:agg}
Given locally updated $\{\theta_k^{t}\}_{k=1}^K$, which are collected at the end of round $t$, the aggregation step of round $t$ can be formulated as $\theta^{t} = \theta_{0}^{t} = \sum_{k = 1}^K a_{k} \theta_{k}^{t}$, where $a_k$ denotes $k^{\mathrm{th}}$ element of diagonal matrix $A$. For example, in \texttt{FedAvg} \cite{mcmahan2017communication}, we use $a_k = \frac{n_k}{\sum_{j = 1}^K n_j}$ because the number of labels indicates the strength of supervision in each node, which does not work for unsupervised learning. For node $k$ with large $n_k$ and small variation in $\mathcal{D}_k$ (i.e.~examples are similar), $f_{\theta_k}$ converges faster but $\theta_k$ learns less meaningful representations (e.g.~overfitted to no findings or to certain diseases). But $a_k = \frac{n_k}{\sum_{j = 1}^K n_j}$ gives $\theta_k^t$ a larger weight and $\theta_k^0$ is dominated by $\theta_k^t$. Instead, we propose \textit{self-adaptive aggregation} to compute matrix $A$.

\begin{algorithm}[t]
    \begin{algorithmic}
    \State{Initialize $\theta_0^0$} 
    \For{$t = 1, 2, \cdots, T$}
    \For{$k = 1, 2, \cdots, K$}
    \State{$\theta_k^{t} \leftarrow \theta_0^{t-1}$} \Comment{Synchronize with PS}
    \If{$t \leq t_w$} \Comment{Warm up}
    \State{$\theta_k^{t} \leftarrow$ \textit{local\_update}$(\theta_k^{t})$} \Comment{Eq.~\ref{eq:nce}}
    \Else \Comment{Metadata transfer}
    \State{Download $\{(\bm{\mu}_j, \bm{\Sigma}_j)\}_{j \neq k}$ from PS}
    \State{$\theta_k^{t} \leftarrow$ \textit{local\_update}$(\theta_k^{t}, \{(\bm{\mu}_j, \bm{\Sigma}_j)\}_{j \neq k})$}
     \Comment{Eq.~\ref{eq:meta_trans}}
    \State{Upload $(\bm{\mu}_k, \bm{\Sigma}_k)$} to PS
    \EndIf
    \EndFor
    \State{Compute $A$} \Comment{Eq.~\ref{eq:rsa} and Eq.~\ref{eq:a}}
    \State{$\theta_0^{t} \leftarrow \sum_{k = 1}^K a_{k}^t \theta_{k}^{t}$}
    \EndFor
    \State{Output $\theta_0^T$}
    \end{algorithmic}
    \caption{\footnotesize \texttt{FedMoCo}. The training in each data node is warmed up for $t_w$ rounds before metadata transfer. A local round could be a few local epochs.}
    \label{algo:1}
\end{algorithm}

Let $-1 \leq r_k \leq 1$ denote representational similarity analysis (RSA) \cite{kriegeskorte2008representational} of node $k$. In round $t$ we first take $n_k{'}$ random samples from $\mathcal{D}_k$ as a subset $\mathcal{D}_k^{'}$ for computational and statistical efficiency. For each $x_i \in \mathcal{D}_k^{'}$, we get two representations $f_{\theta_0^{t-1}}(x_i)$ (aggregated weights at the end of round $t-1$, which are also globally synchronized weights at the beginning of round $t$) and $f_{\theta_k^{t}}(x_i)$ (locally updated weights at the end of round $t$). We define $\rho_{ij}$ as Pearson's correlation coefficient between $f_{\theta}(x_i)$ and $f_{\theta}(x_j)$ $\forall \ 0 \leq i,j \leq n_k{'}$ and define $\mathrm{RDM}_{ij} = 1 - \rho_{ij}$, where RDM is representation dissimilarity matrix \cite{kriegeskorte2008representational} for $f_{\theta}$. Then, $r_k$ is defined based on Spearman's rank correlation:
\begin{equation}
    r_k = 1 - \frac{6 \sum_{i=1}^n d_i^2}{n(n^2 -1)}
    \label{eq:rsa}
\end{equation}
where $d_i$ is the difference between the ranks of $i^{\mathrm{th}}$ elements of the lower triangular of $\mathrm{RDM}$ for $f_{\theta^{t-1}}$ and $\mathrm{RDM}$ for $f_{\theta_k^{t}}$, and $n = n_k{'}(n_k{'} - 1) / 2$. We define $A$ for the aggregation step at the end of round $t$ as
\begin{equation}
   a_k^t = \frac{1 - r_k}{\sum_{j = 1}^K 1 - r_j}.
   \label{eq:a}
\end{equation} 
The CL performance at node $k$ in round $t$ is measured by $r_k$. 
Intuitively, given the same $\theta^{t-1}$ at the start of round $t$, a small $r_k$ indicates that the change between the representations $f_{\theta^{t-1}}$ and $f_{\theta^{t}}$ is large, i.e. node $k$ is still learning meaningful representations. Eq.~\ref{eq:a} assigns larger weights to local models with higher potentials in representational power. The complete pseudo-code is given in Algorithm~\ref{algo:1}.

\section{Experiments}
\subsection{Experimental Setup}
\label{sec:exp_setup}
For a fair comparison, we use the same set of hyperparameters and the same training strategy for all experiments. We use ResNet18 \cite{he2016deep} as the network backbone, initialized with the same random seed. We use the Momentum optimizer with momentum $0.9$. Without using any pre-trained weights, we demonstrate that \texttt{FedMoCo} can learn from scratch. The initial learning rate is $0.03$ and is multiplied by $0.1$ (and $0.01$) at $120$ (and $160$) epochs. The batch size is $64$ for each node, the weight decay is $10^{-4}$, $m$ in Eq.~\ref{eq:moment} is $0.999$, $\lambda$ in Eq.~\ref{eq:bc} is $0.5$, $\tau$ is $0.2$, $N$ is 1024, and $\eta$ in Eq.~\ref{eq:meta_trans} is $0.05$. We estimate RSA with $n_k{'} = 100$.

In the absence of FCL for MIA, we compare \texttt{FedMoCo} with a strong baseline which is an integration of MoCo \cite{he2020momentum} and \texttt{FedAvg} \cite{mcmahan2017communication}, a seminal supervised FL model. All models are implemented with PyTorch on NVIDIA Tesla V100.

\begin{figure}[t]
    \centering
    \begin{subfigure}[t]{0.16\textwidth}
    \includegraphics[width=\textwidth]{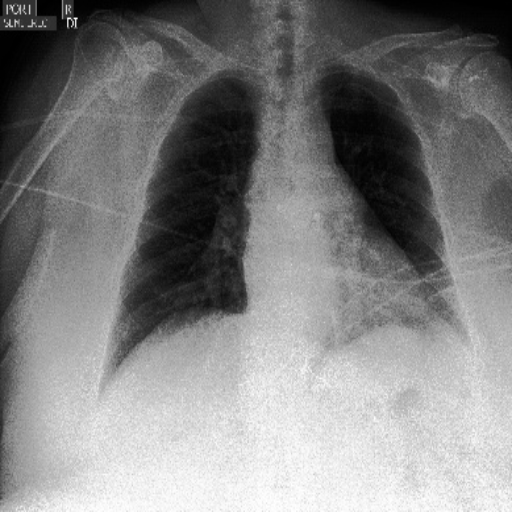}
    \caption{}
    \end{subfigure}
    \begin{subfigure}[t]{0.16\textwidth}
    \includegraphics[width=\textwidth]{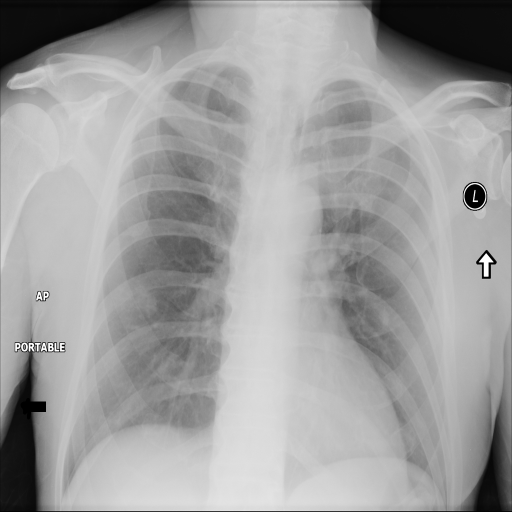}
    \caption{}
    \end{subfigure}
    \begin{subfigure}[t]{0.16\textwidth}
    \includegraphics[width=\textwidth]{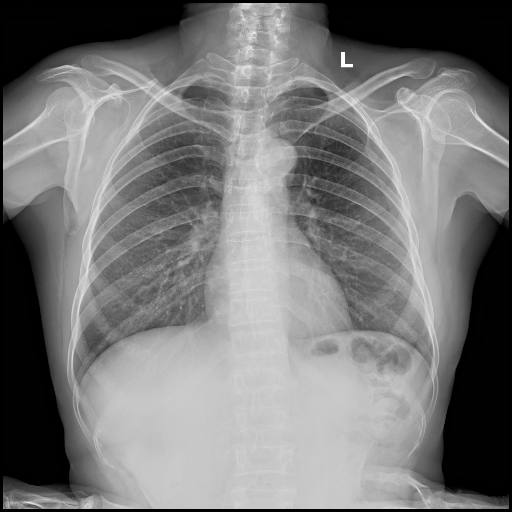}
    \caption{}
    \end{subfigure}
    \caption{Pre-training on: (a) CheXpert; (b) ChestX-ray8; (c) VinDr-CXR.}
    \label{fig:data}
\end{figure}

\begin{figure}[t]
    \centering
    \captionsetup{justification=centering}
    \begin{subfigure}[t]{0.18\textwidth}
    \includegraphics[width=\textwidth]{figs/raw.png}
    \caption{}
    \end{subfigure}
    \begin{subfigure}[t]{0.18\textwidth}
    \includegraphics[width=\textwidth]{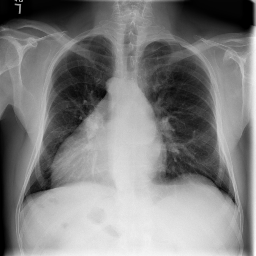}
    \caption{}
    \end{subfigure}
    \begin{subfigure}[t]{0.18\textwidth}
    \includegraphics[width=\textwidth]{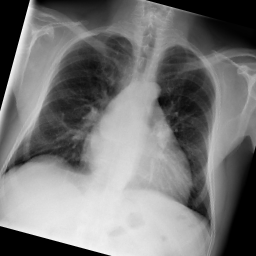}
    \caption{}
    \end{subfigure}
    \begin{subfigure}[t]{0.18\textwidth}
    \includegraphics[width=\textwidth]{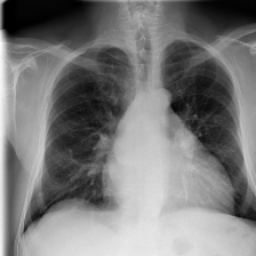}
    \caption{}
    \end{subfigure}
    \begin{subfigure}[t]{0.18\textwidth}
    \includegraphics[width=\textwidth]{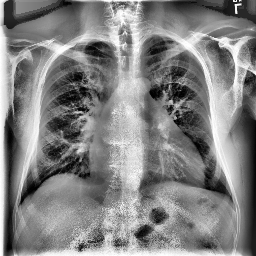}
    \caption{}
    \end{subfigure}
    \caption{Stochastic data augmentation: (a) raw; (b) flip; (c) rotate; (d) crop and resize; (e) histogram equalization.}
    \label{fig:aug}
\end{figure}

\begin{table}[t]
    \centering
    \captionsetup{justification=centering}
    {\footnotesize
    \begin{tabular}{l | c | c | c | c | c | c}
        dataset  &  size & \# of classes & multi-label & multi-view & balanced & resolution \\\Xhline{4\arrayrulewidth}
        CheXpert \cite{irvin2019chexpert} & 371920  & 14 & \cmark & \cmark & \xmark & $390 \times 320$\\\hline
        ChestX-ray8 \cite{wang2017chestx} & 112120 & 15 & \cmark & \xmark  & \xmark & $ 1024 \times 1024 $\\\hline
        VinDr-CXR \cite{nguyen2020vindr} & 15000 & 15 & \cmark & \xmark  & \xmark & $512 \times 512$\\\hline
        COVID-19 \cite{chowdhury2020can} & 3886 & 3 & \xmark & \xmark & \cmark & $256 \times 256$ \\\hline
    \end{tabular}
    }
    \caption{Dataset description.}
    \label{tab:data}
\end{table}

\textbf{Datasets.} FCL should work for any type of medical images.
We illustrate FCL on anterior and posterior chest X-rays (CXRs). We use three public large-scale CXR datasets as the unlabeled pre-training data to simulate the federated environment, namely CheXpert \cite{irvin2019chexpert}, ChestX-ray8 \cite{wang2017chestx}, and VinDr-CXR \cite{nguyen2020vindr} (see Table~\ref{tab:data}). Three datasets are collected and annotated from different sources independently and express a large variety in data modalities (see Fig.~\ref{fig:data}). The original images are cropped and downsampled to $256 \times 256$. Three datasets contain noisy labels and the label distributions are different.

\textbf{Data Augmentation.}
Driven by clinical domain knowledge and experimental findings, We provide a augmentation policy as illustrated in Fig.~\ref{fig:aug}. Compared with previous studies of MoCo on medical images \cite{he2020sample,chen2020momentum,sowrirajan2020moco}, we propose stochastic histogram equalization for CL on medical images. The stochasticity comes from uniformly sampling the parameters of CLAHE \cite{pizer1987adaptive}. For a fair comparison, we use the same augmentation policy for all models in this work.

\textbf{Linear Classification Protocol.}
We evaluate the performance of unsupervised pre-training by following \textit{linear classification protocol}(LCP) \cite{he2020momentum,misra2020self,chen2020simple,tian2020makes}. ResNet18 is first pre-trained on the unlabeled dataset. Then a supervised linear classifier (a fully-connected layer) is trained on top of the frozen features extracted from a labeled dataset for $50$ epochs with a constant learning rate $0.1$. We report the classification accuracy on the validation set of the labeled data as the performance of FCL. We choose a public COVID-19 CXR dataset \cite{chowdhury2020can} with 3886 CXRs. Note that COVID-19 has not been seen in three pre-training datasets. 
We use $50\%$ CXRs for training and $50\%$ for testing.

\subsection{Evaluation}
The experiments are simulated in a controllable federated environment. To eliminate the effect of $n_k$, we first evaluate \texttt{FedMoCo} in a situation that each node has the same number of CXRs. We create 3 data nodes by randomly sampling $10000$ CXRs from each of three datasets. We create 6 nodes by partitioning each data node equally. We evaluate the pre-training performance on $T = 200$ and $T = 400$, where one round is one local epoch and \texttt{FedMoCo} is warmed up for $t_w = 50$ epochs. We provide the performance of MoCo trained in a single node with the same data as \textit{Oracle} for centralized CL. LCP results with mean and standard deviation of 3 rounds are present in Table~\ref{tab:1}. We have two empirical findings for FCL: more nodes will decrease the performance, and more training time may not improve the performance of downstream tasks.

To show the isolated contribution of metadata transfer and self-adaptive aggregation, we simulate two common situations in FL with $K = 3$ and $T = 200$. First, with the same data as above, we create unbalanced data distributions in terms of the number of examples by only using $\gamma \%$ CXRs in nodes of CheXpert and ChestX-ray8. We use \texttt{FedMoCo-M} to denote \texttt{FedMoCo} with only metadata transfer module and use \texttt{FedMoCo-S} to denote \texttt{FedMoCo} with only self-adaptive aggregation. The results are presented in Table~\ref{tab:2}. Second, with the same number of examples in each node, we create unbalanced data distributions in terms of the label distribution by only sampling healthy CXRs (no findings) for the nodes of CheXpert and ChestX-ray8 and only sampling CXRs with disease labels for the node of VinDr-CXR. The results are presented in Table~\ref{tab:3}. In summary the above results, \texttt{FedMoCo} can outperform \texttt{FedAvg} consistently under non-IID challenges while metadata transfer and self-adaptive aggregation both can improve the performance as an individual module. An interesting finding against common sense is that sometimes FCL can outperform centralized CL depending on the data distributions, which may worth further investigation.

\begin{table}[t]
    \begin{minipage}[t]{0.49\textwidth}
        \centering
        \captionsetup{justification=centering}
        {\footnotesize
        \begin{tabular}{l|c|c|c}
            Model  & K & $T = 200$ ($\%$) & $T = 400$ ($\%$) \\\Xhline{4\arrayrulewidth}
            \texttt{FedAvg} & 3 & 94.92 $\pm$ 0.16 & 95.03 $\pm$ 0.23 \\\hline
            \texttt{FedMoCo} & 3 & 95.73 $\pm$ 0.05 & 95.58 $\pm$ 0.26\\\hline
            \texttt{FedAvg} & 6 & 94.60 $\pm$ 0.32 & 94.65 $\pm$ 0.94 \\\hline
            \texttt{FedMoCo} & 6 & 95.11 $\pm$ 0.71 & 94.92 $\pm$ 1.27 \\\Xhline{3\arrayrulewidth}
            \textit{Oracle} & 1 & 92.80 $\pm$ 0.05 & 92.37 $\pm$ 0.06 \\\hline
        \end{tabular}
        }
        \caption{LCP comparison between FCL models on non-IID distributions with equal numbers of examples.}
        \label{tab:1}
    \end{minipage}
    \begin{minipage}[t]{0.49\textwidth}
        \centering
        \captionsetup{justification=centering}
        {\footnotesize
        \begin{tabular}{l | c | c}
            Model  & $\gamma = 5$ ($\%$) & $\gamma = 10$ ($\%$) \\\Xhline{4\arrayrulewidth}
            \texttt{FedAvg} & 94.27 $\pm$ 1.37 & 94.77 $\pm$ 0.86 \\\hline
            \texttt{FedMoCo-M} & 94.66 $\pm$ 1.57 & 95.23 $\pm$ 0.96 \\\hline
            \texttt{FedMoCo-S} & 94.37 $\pm$ 0.77 & 95.01 $\pm$ 0.45\\\hline
            \texttt{FedMoCo} & 94.77 $\pm$ 0.80 & 95.58 $\pm$ 0.31 \\\Xhline{3\arrayrulewidth}
            \textit{Oracle} & 95.22 $\pm$ 0.05 & 95.23 $\pm$ 0.39 \\\hline
        \end{tabular}
        }
        \caption{LCP comparison between FCL models on non-IID distributions with imbalanced numbers of examples.}
        \label{tab:2}
    \end{minipage}
\end{table}

\begin{table}[t]
    \begin{minipage}[t]{0.53\textwidth}
        \centering
        \captionsetup{justification=centering}
        {\footnotesize
        \begin{tabular}{l | c | c}
            Model  & $n_k=5000$ ($\%$) & $n_k=10000$ ($\%$) \\\Xhline{4\arrayrulewidth}
            \texttt{FedAvg} & 93.30 $\pm$ 1.21 & 95.25 $\pm$ 0.33 \\\hline
            \texttt{FedMoCo-M} & 93.67 $\pm$ 1.19 & 95.43 $\pm$ 0.36\\\hline
            \texttt{FedMoCo-S} & 94.11 $\pm$ 0.88 & 95.74 $\pm$ 0.29\\\hline
            \texttt{FedMoCo} & 94.38 $\pm$ 0.68 & 96.02 $\pm$ 0.25 \\\Xhline{3\arrayrulewidth}
            \textit{Oracle} & 95.78 $\pm$ 0.22 & 94.82 $\pm$ 0.11 \\\hline
        \end{tabular}
        }
        \caption{LCP comparison between FCL models on non-IID distributions with imbalanced label distribution.}
        \label{tab:3}
    \end{minipage}
    \begin{minipage}[t]{0.45\textwidth}
        \centering
        \captionsetup{justification=centering}
        {\footnotesize
        \begin{tabular}{l | c }
            Model & Accuracy ($\%$) \\\Xhline{4\arrayrulewidth}
            w/o Pre-Training & 79.63 \\\hline
            \texttt{FedAvg} & 87.24 \\\hline
            \texttt{FedMoCo} & 91.56 \\\Xhline{3\arrayrulewidth}
            \textit{Oracle}(MoCo) &  88.42 \\\hline
            \textit{Oracle}(Supervised) & 95.78 \\\hline
        \end{tabular}
        }
        \caption{Fine-tuning models pre-trained by FCL for COVID-19 detection.}
        \label{tab:4}
    \end{minipage}
\end{table}

\textbf{Downstream Task Evaluation.} To demonstrate the practical value of the representations extracted by FCL under data scarcity, we use COVID-19 detection \cite{chowdhury2020can} as the downstream task. We use the same training set and test set in LCP. We fine-tune the ResNets pre-trained by FCL models in Table~\ref{tab:1} ($K = 3$ and $T = 200$) with only $3\%$ of the training set. We train a randomly initialized ResNet with the full training set as the \textit{Oracle} for supervised learning. For a fair comparison, we use a fixed learning rate $0.01$ to  train all models for 100 epochs. We report the highest accuracy in Table~\ref{tab:4}. \texttt{FedMoCo} outperforms centralized CL and \texttt{FedAvg}. Compared with standard supervised learning, \texttt{FedMoCo} utilizes only $3\%$ of labels to achieve $90\%$ accuracy, which greatly reduces the annotation cost.

\section{Conclusion}
In this work, we formulate and discuss FCL on medical images and propose \texttt{FedMoCo}. We evaluate the robustness of \texttt{FedMoCo} under a few characteristic non-IID challenges and use COVID-19 detection as the downstream task. More investigations will be conducted, but these initial results already provide insights into future FCL research. In future, we plan to focus on the task affinity \cite{vandenhende2020branched} between FCL and its corresponding downstream tasks. We will quantitatively analyze how the representations extracted by FCL can influence the performance of different downstream tasks.

\subsubsection*{Acknowledgements}
We would like to thank Huawei Technologies Co., Ltd. for providing GPU computing service for this study.
%
%
%
\bibliographystyle{splncs04}
\bibliography{ref}

\begin{thebibliography}{10}
\providecommand{\url}[1]{\texttt{#1}}
\providecommand{\urlprefix}{URL }
\providecommand{\doi}[1]{https://doi.org/#1}

\bibitem{baldi1991contrastive}
Baldi, P., Pineda, F.: Contrastive learning and neural oscillations. Neural
  Computation  \textbf{3}(4),  526--545 (1991)

\bibitem{box1964analysis}
Box, G.E., Cox, D.R.: An analysis of transformations. Journal of the Royal
  Statistical Society: Series B (Methodological)  \textbf{26}(2),  211--243
  (1964)

\bibitem{chen2020simple}
Chen, T., Kornblith, S., Norouzi, M., Hinton, G.: A simple framework for
  contrastive learning of visual representations. In: ICML (2020)

\bibitem{chen2020momentum}
Chen, X., Yao, L., Zhou, T., Dong, J., Zhang, Y.: Momentum contrastive learning
  for few-shot covid-19 diagnosis from chest ct images. Pattern Recognition p.
  107826 (2020)

\bibitem{chowdhury2020can}
Chowdhury, M.E., Rahman, T., Khandakar, A., Mazhar, R., Kadir, M.A., Mahbub,
  Z.B., Islam, K.R., Khan, M.S., Iqbal, A., Al~Emadi, N., et~al.: Can ai help
  in screening viral and covid-19 pneumonia? IEEE Access  \textbf{8},
  132665--132676 (2020)

\bibitem{dong2018unsupervised}
Dong, N., Kampffmeyer, M., Liang, X., Wang, Z., Dai, W., Xing, E.: Unsupervised
  domain adaptation for automatic estimation of cardiothoracic ratio. In:
  MICCAI. pp. 544--552. Springer (2018)

\bibitem{dou2021federated}
Dou, Q., So, T.Y., Jiang, M., Liu, Q., Vardhanabhuti, V., Kaissis, G., Li, Z.,
  Si, W., Lee, H.H., Yu, K., et~al.: Federated deep learning for detecting
  covid-19 lung abnormalities in ct: a privacy-preserving multinational
  validation study. NPJ digital medicine  \textbf{4}(1),  1--11 (2021)

\bibitem{he2020momentum}
He, K., Fan, H., Wu, Y., Xie, S., Girshick, R.: Momentum contrast for
  unsupervised visual representation learning. In: CVPR. pp. 9729--9738 (2020)

\bibitem{he2016deep}
He, K., Zhang, X., Ren, S., Sun, J.: Deep residual learning for image
  recognition. In: CVPR. pp. 770--778 (2016)

\bibitem{he2020sample}
He, X., Yang, X., Zhang, S., Zhao, J., Zhang, Y., Xing, E., Xie, P.:
  Sample-efficient deep learning for covid-19 diagnosis based on ct scans.
  MedRxiv  (2020)

\bibitem{irvin2019chexpert}
Irvin, J., Rajpurkar, P., Ko, M., Yu, Y., Ciurea-Ilcus, S., Chute, C.,
  Marklund, H., Haghgoo, B., Ball, R., Shpanskaya, K., et~al.: Chexpert: A
  large chest radiograph dataset with uncertainty labels and expert comparison.
  In: AAAI. vol. 33(01), pp. 590--597 (2019)

\bibitem{kriegeskorte2008representational}
Kriegeskorte, N., Mur, M., Bandettini, P.A.: Representational similarity
  analysis-connecting the branches of systems neuroscience. Frontiers in
  systems neuroscience  \textbf{2}, ~4 (2008)

\bibitem{li2014communication}
Li, M., Andersen, D.G., Smola, A.J., Yu, K.: Communication efficient
  distributed machine learning with the parameter server. In: NIPS. pp. 19--27
  (2014)

\bibitem{li2019privacy}
Li, W., Milletar{\`\i}, F., Xu, D., Rieke, N., Hancox, J., Zhu, W., Baust, M.,
  Cheng, Y., Ourselin, S., Cardoso, M.J., et~al.: Privacy-preserving federated
  brain tumour segmentation. In: International Workshop on Machine Learning in
  Medical Imaging. pp. 133--141. Springer (2019)

\bibitem{mcmahan2017communication}
McMahan, B., Moore, E., Ramage, D., Hampson, S., Aguera~y Arcas, B.:
  Communication-efficient learning of deep networks from decentralized data.
  In: AISTATS. pp. 1273--1282. PMLR (2017)

\bibitem{misra2020self}
Misra, I., Maaten, L.v.d.: Self-supervised learning of pretext-invariant
  representations. In: CVPR. pp. 6707--6717 (2020)

\bibitem{nguyen2020vindr}
Nguyen, H.Q., Lam, K., Le, L.T., Pham, H.H., Tran, D.Q., Nguyen, D.B., Le,
  D.D., Pham, C.M., Tong, H.T., Dinh, D.H., et~al.: Vindr-cxr: An open dataset
  of chest x-rays with radiologist's annotations. arXiv preprint
  arXiv:2012.15029  (2020)

\bibitem{oord2018representation}
Oord, A.v.d., Li, Y., Vinyals, O.: Representation learning with contrastive
  predictive coding. arXiv preprint arXiv:1807.03748  (2018)

\bibitem{pizer1987adaptive}
Pizer, S.M., Amburn, E.P., Austin, J.D., Cromartie, R., Geselowitz, A., Greer,
  T., ter Haar~Romeny, B., Zimmerman, J.B., Zuiderveld, K.: Adaptive histogram
  equalization and its variations. Computer vision, graphics, and image
  processing  \textbf{39}(3),  355--368 (1987)

\bibitem{rieke2020future}
Rieke, N., Hancox, J., Li, W., Milletari, F., Roth, H.R., Albarqouni, S.,
  Bakas, S., Galtier, M.N., Landman, B.A., Maier-Hein, K., et~al.: The future
  of digital health with federated learning. NPJ digital medicine
  \textbf{3}(1), ~1--7 (2020)

\bibitem{de1994learning}
de~Sa, V.R.: Learning classification with unlabeled data. In: NIPS. pp.
  112--119. Citeseer (1994)

\bibitem{sheller2020federated}
Sheller, M.J., Edwards, B., Reina, G.A., Martin, J., Pati, S., Kotrotsou, A.,
  Milchenko, M., Xu, W., Marcus, D., Colen, R.R., et~al.: Federated learning in
  medicine: facilitating multi-institutional collaborations without sharing
  patient data. Scientific Reports  \textbf{10}(1),  1--12 (2020)

\bibitem{sowrirajan2020moco}
Sowrirajan, H., Yang, J., Ng, A.Y., Rajpurkar, P.: {MoCo} pretraining improves
  representation and transferability of chest x-ray models. arXiv preprint
  arXiv:2010.05352  (2020)

\bibitem{tian2020makes}
Tian, Y., Sun, C., Poole, B., Krishnan, D., Schmid, C., Isola, P.: What makes
  for good views for contrastive learning. In: NIPS (2020)

\bibitem{vandenhende2020branched}
Vandenhende, S., Georgoulis, S., De~Brabandere, B., Van~Gool, L.: Branched
  multi-task networks: deciding what layers to share. In: BMVC (2020)

\bibitem{wang2017chestx}
Wang, X., Peng, Y., Lu, L., Lu, Z., Bagheri, M., Summers, R.M.: Chestx-ray8:
  Hospital-scale chest x-ray database and benchmarks on weakly-supervised
  classification and localization of common thorax diseases. In: CVPR. pp.
  2097--2106 (2017)

\bibitem{wu2018unsupervised}
Wu, Z., Xiong, Y., Stella, X.Y., Lin, D.: Unsupervised feature learning via
  non-parametric instance discrimination. In: CVPR (2018)

\end{thebibliography}
\end{document}